# DeepVesselNet: Vessel Segmentation, Centerline Prediction, and Bifurcation Detection in 3-D Angiographic Volumes

Giles Tetteh*, Velizar Efremov, Nils D. Forkert, Matthias Schneider, Jan Kirschke, Bruno Weber, Claus Zimmer, Marie Piraud, and Björn H. Menze,

*Abstract*—We present DeepVesselNet, an architecture tailored to the challenges faced when extracting vessel networks or trees and corresponding features in 3-D angiographic volumes using deep learning. We discuss the problems of low execution speed and high memory requirements associated with full 3-D convolutional networks, high-class imbalance arising from the low percentage (less than 3%) of vessel voxels, and unavailability of accurately annotated training data - and offer solutions as the building blocks of DeepVesselNet.

First, we formulate 2-D orthogonal cross-hair filters which make use of 3-D context information at a reduced computational burden. Second, we introduce a class balancing cross-entropy loss function with false positive rate correction to handle the high-class imbalance and high false positive rate problems associated with existing loss functions. Finally, we generate synthetic dataset using a computational angiogenesis model capable of generating vascular trees under physiological constraints on local network structure and topology and use these data for transfer learning. DeepVesselNet is optimized for segmenting and analyzing vessels, and we test the performance on a range of angiographic volumes including clinical time-of-flight MRA data of the human brain, as well as synchrotron radiation X-ray tomographic microscopy scans of the rat brain. Our experiments show that, by replacing 3-D filters with cross-hair filters in our network, we achieve over 23% improvement in speed, lower memory footprint, lower network complexity which prevents overfitting and comparable accuracy (with a Cox-Wilcoxon paired sample significance test p-value of 0.07 when compared to full 3-D filters). Our class balancing metric is crucial for training the network and transfer learning with synthetic data is an efficient, robust, and very generalizable approach leading to a network that excels in a variety of angiography segmentation tasks. We show that sub-sampling and max pooling layers may lead to drop in performance in tasks that involve voxel-sized structures and DeepVesselNet architecture which does not use any form of sub-sampling layer works well for vessel segmentation, centerline prediction and bifurcation detection.

We make available our datasets publicly for downloading, fostering future research, in particular, on centerline prediction and bifurcation detection and serving as one of the first public datasets for brain vessel tree segmentation and analysis.

*G. Tetteh, V. Efremov, M. Piraud, and B.H. Menze are with the Department of Computer Science, TU München, München, Germany.

J. Kirschke and C. Zimmer are with the Neuroradiology, Klinikum Rechts der Isar, TU München, München, Germany.

N. D. Forkert is with the Department of Radiology, University of Calgary, Calgary, Canada.

M. Schneider and B. Weber are with the Institute of Pharmacology and Toxicology, University of Zurich, Zurich, Switzerland.

*Asterisk indicates corresponding author*

## I. INTRODUCTION

Angiography offers insights into blood flow and conditions of the vascular tree. Three dimensional volumetric angiography information can be obtained using magnetic resonance (MRA), ultrasound, or x-ray based technologies like computed tomography (CT). A common first step in analyzing these data is vessel segmentation. Still, moving from raw angiography images to vessel segmentation alone might not provide enough information for clinical use, and other vessel features like centerline, diameter, or bifurcations of the vessels are also needed to accurately extract information about the vascular tree, for example, to characterize its structural properties or flow pattern. In this work, we present a deep learning approach, called DeepVesselNet, to perform vessel segmentation, centerline prediction, and bifurcation detection tasks. DeepVesselNet deals with challenges that result from speed and memory requirements, unbalanced class labels, and the difficulty of obtaining well-annotated data for curvilinear volumetric structures by addressing the following three key limitations.

*a) Fast cross-hair filters:* Processing 3-D medical volumes poses a memory consumption and speed challenge. Using 3-D CNNs leads to drastic increase in number of parameters to be optimized and computations to be executed when compared to 2-D CNNs. At the same time, applying a 2-D CNN in a slice-wise fashion discards valuable 3-D context information that is crucial for tracking curvilinear structures in 3-D. Inspired by the ideas of [1]–[3] who proposed separable filters and using intersecting 2-D planes, we demonstrate the use of cross-hair filters from three intersecting 2-D filters, which helps to avoid the memory and speed problems of classical 3-D networks, while at the same time making use of 3-D information in volumetric data. Unlike the existing ideas where 2-D planes are extracted at a pre-processing stage and used as input channels (see discussion in Sec. III-A0b), our cross-hair filters (see Sec. III-A) are implemented on a layer level which help to retain the 3-D information throughout the network.

*b) Extreme class balancing:* The vessel, centerline and bifurcation prediction tasks is characterized by high class imbalances. Vessels account for less than 3% of the total voxels in a patient volume, centerlines represent a fraction of the segmented vessels, and visible bifurcations are in the hundreds at best – even when dealing with volumes with $10^6$ and more voxels. This bias towards the background class is a common problem in medical data [4]–[6]. Unfortunately, current class balancing loss functions for training CNNs turn out to be numerically unstable in extreme cases as ours. We offer a solution to this 'extreme class imbalance' problem by introducing a new loss function (see Sec. III-B) that we demonstrate to work well with our vascular features of interest.

*c) Transfer learning from synthetic data:* Manually annotating vessels, centerlines, and bifurcations requires many hours of work and expertise. To this end, we demonstrate the successful use of simulation based frameworks [7]–[9] that can be used for generating synthetic data with accurate labels (see Sec. III-C) for pretraining our networks, rendering the training of our supervised classification algorithm feasible. The transfer learning approach turns out to be a critical component for training CNNs in a wide range of angiography tasks and applications ranging from CT micrographs to TOF MRA. The synthesized and the clinical MRA datasets are made available publicly for future research and validation purposes. Further description and download link is provided in Section IV-A.

## II. PRIOR WORK AND OPEN CHALLENGES

*a) Vessel Segmentation:* Vessel enhancement and segmentation is a longstanding task in medical image analysis (see reviews by [10], [11]). The range of methods employed for vessel segmentation reflect the development of image processing during the past decades, including region growing techniques [12], active contours [13], statistical and shape models [14]–[17], particle filtering [18]–[20] and path tracing [21]. All of these examples are interactive, starting from a set of seed labels as root and propagating towards the branches. Other approaches aim at an unsupervised enhancement of vascular structures: a popular multi-scale second order local structure of an image (Hessian) was examined by [22] with the purpose of developing a vessel enhancement filter. A measure of vessel-likeliness is then obtained as a function of all eigenvalues of the Hessian. A novel curvilinear structure detector, called Optimally Oriented Flux (OOF) was proposed by [23] to find an optimal axis on which image gradients are projected to compute the image gradient flux. OOF has a lower computational load than the calculation of the Hessian matrix proposed in [22]. A level-set segmentation approach with vesselness-dependent anisotropic energy weights is presented and evaluated in [24], [25] for 3-D time-of-flight (TOF) MRA. Phellan and Forkert [26] presented a comparative analysis of the accuracy gains in vessel segmentation generated by the use of nine vessel enhancement algorithms on time-of-flight MRA that included multi scale vesselness algorithms, diffusion-based filters, and filters that enhance tubular shapes and concluded that vessel enhancement algorithms do not always lead to more accurate segmentation results compared to segmenting non-enhanced images directly. An early machine learning approach for vessel segmentation was proposed by [27], combining joint 3-D vessel segmentation and centerline extraction using oblique Hough forest with steerable filters. In a similar fashion, [28] used deep artificial neural network as a pixel classifier to automatically segment neuronal structures in stacks of electron microscopy images, a task somewhat similar to vessel segmentation. One example using deep learning architecture is [29] who used a deep convolutional neural network to automatically segment the vessels of the brain in TOF MRA by extracting manually annotated bi-dimensional image patches in the axial, coronal, and sagittal directions as an input to the training process. Koziński et. al. [30] proposed a loss function that accommodates ground truth annotations of 2-D projections of the training volumes, for training deep neural networks in tasks where obtaining full 3-D annotations is a challenge.

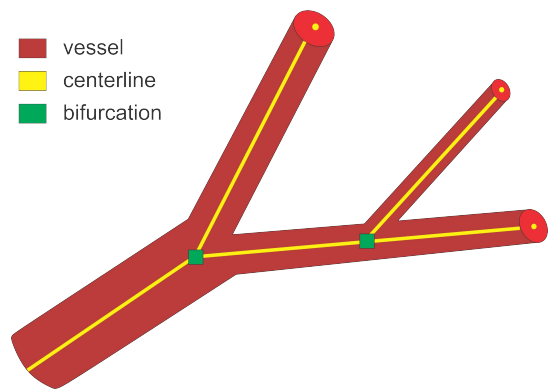

Fig. 1. An overview of the three main tasks tackled in this paper. For bifurcations, we predict a neigbourhood cube around the indicated point.

Though deep learning has been applied in many medical imaging tasks, there are no dedicated architectures so far for vessel segmentation in 3-D volumetric datasets. Existing architectures might be sub-optimal and not work directly out of the box due to the unique nature of the vasculature as compared to other imaging tasks. There is therefore the need to explore other architectures and training strategies.

*b) Centerline Prediction:* Identifying the center of a vessel is relevant for calculating the vessel diameter, but also for obtaining the 'skeleton' of a vessel when extracting the vascular tree or network (see Fig. 1). The vessels' skeleton and center can be found by post-processing a previously generated vessel segmentation. A method based on morphological operations is developed by [31] which performs erosion using $2 \times 2$ neighborhoods of a pixel to determine if a pixel is a centerline candidate. Active contour models are applied in [32] as well as path planning and distance transforms for extracting centerline in vessels, and [33] proposed a geodesic or minimal path technique. A morphology-guided level set model is used in [34] to performed centerline extraction by learning the structural patterns of a tubular-like object, and estimating the centerline of a tubular object as the path with minimal cost with respect to outward flux in gray level images. Vesselness filters were adopted by [35] to predict the location of the centerline, while [36] used Hough transforms in handling the similar task. A Hough random forest with local image filters is designed in [7], [27] to predict the centerline, and trained on centerline data previously extracted using one of the level set approaches.

The application of deep learning to the extraction of vessel centerline has not been explored. One reason may be the lack of annotated data necessary to train deep architectures that is hard to obtain especially in 3-D datasets.

*c) Bifurcation Detection:* A vessel bifurcation refers to the point on a vessel centerline where the vessel splits into two vessels (see Fig. 1). Bifurcations represent the nodes of the vascular tree or network and knowing their locations is important both for network extraction and for studying its properties [37]. They represent structures that can easily be used as landmarks in image registration, but also indicate the locations of modified blood flow velocity and pressure within the network itself [38]. Bifurcations are hard to detect in volumetric data as they are rare point-like features that vary in size and shape significantly. Similar to centerline extraction, the detection of bifurcations often happens by postprocessing a previously generated vessels segmentation or by searching a previously extracted vessel graph. A two staged deep learning architecture is proposed in [39] for detecting carotid artery bifurcations as a specific landmark in volumetric CT data by first training a shallow network for predicting candidate regions followed by a sparse deep network for final prediction. A three stage algorithm for bifurcation detection is proposed in [38] for digital eye fundus images, a 2-D task, and their approach included image enhancement, clustering, and searching the graph for bifurcations.

The direct predicting of the location of centerlines and bifurcations without a previous segmentation of vessels as an intermediate step is a task which has not been attempted yet. We foresee that having directly predicted centerlines and bifurcations together with those from postprocessing vessel segmentations will enhance the overall robustness and accuracy of the analysis of angiographic volumes.

## III. METHODOLOGY

### A. Cross-hair Filters Formulation

In this section, we introduce the 3-D convolutional operator, which utilizes cross-hair filters to improve speed and memory usage while maintaining accuracy. Let $I$ be a 3-D volume, M a 3-D convolutional kernel of shape $(k_x, k_y, k_z)$, and $*$ be a convolutional operator. We define $*$ as:

$$I * M = A = \{a_{ijk}\}; \quad a_{ijk} = \sum_{r=1}^{k_x}\sum_{s=1}^{k_y}\sum_{t=1}^{k_z} I_{(R,S,T)} M_{(r,s,t)}; \quad (1)$$
$$R = i + r - \left(1 + \left[\frac{k_x}{2}\right]\right), \quad S = j + s - \left(1 + \left[\frac{k_y}{2}\right]\right),$$
$$T = k + t - \left(1 + \left[\frac{k_z}{2}\right]\right),$$

where $\{a_{ijk}\}$ is a position element of matrix $A$, $I_{(R,S,T)}$ is the intensity value of image $I$ at voxel position $(R, S, T)$, $M_{(r,s,t)}$ is the value of kernel $M$ at position $(r, s, t)$, and $[x]$ is the greatest integer less or equal to $x$.

From Eq. (1), we see that a classical 3-D convolution involves $k_x k_y k_z$ multiplications and $k_x k_y k_z - 1$ additions for each voxel of the resulting image. For a $3 \times 3 \times 3$ kernel, we have 27 multiplications and 26 additions per voxel. Changing the kernel size to $5 \times 5 \times 5$ increases the complexity to 125 multiplications and 124 additions per voxel. This then scales up with the dimension of the input image. For example, a volume of size $128 \times 128 \times 128$ and a $5 \times 5 \times 5$ kernel results in about $262 \times 10^6$ multiplications and $260 \times 10^6$ additions. To handle this increased computational complexity, we approximate the standard 3-D convolution operation by

$$\begin{aligned} a_{ijk} &= \sum_{s=1}^{k_y}\sum_{t=1}^{k_z} I_{(i,S,T)} M^i_{(s,t)} + \sum_{r=1}^{k_x}\sum_{t=1}^{k_z} I_{(R,j,T)} M^j_{(r,t)} \quad (2) \\ &+ \sum_{r=1}^{k_x}\sum_{s=1}^{k_y} I_{(R,S,k)} M^k_{(r,s)}, \end{aligned}$$

where $M^i, M^j, M^k$ are 2-D convolutional (cross-hair) kernels used as an approximation to the 3-D kernel $M$ in (1) along the $i$th, $j$th, and $k$th axes respectively. $R, S,$ and $T$ are as defined in (1). Using cross-hair filters results in $(k_y k_z + k_x k_z + k_x k_y)$ multiplications and $(k_y k_z + k_x k_z + k_x k_y - 1)$ additions. If we let $k_{m1}, k_{m2}, k_{m3}$ be the sizes of the kernel $M$ such that $k_{m1} \geq k_{m2} \geq k_{m3}$, we can show that

$$k_y k_z + k_x k_z + k_x k_y \leq 3(k_{m1} k_{m2}) \leq k_x k_y k_z, \quad (3)$$

where strict inequality holds for all $k_{m3} > 3$. Eq. 3 shows a better scaling in speed and also in memory since the filters sizes in (1) and (2) are affected by the same inequality. With the approximation in (2), and using the same example as above (volume of size $128 \times 128 \times 128$ and a $5 \times 5 \times 5$ kernel), we now need less than $158 \times 10^6$ multiplications and $156 \times 10^6$ additions to compute the convolution leading to a reduction in computation by more than $100 \times 10^6$ multiplications and additions when compared to a classical 3-D convolution. Increasing the volume or kernel size, further increases the gap between the computational complexity of (1) and (2). Moreover, we will see later from our experiments that (2) still retains essential 3-D context information needed for the classification task.

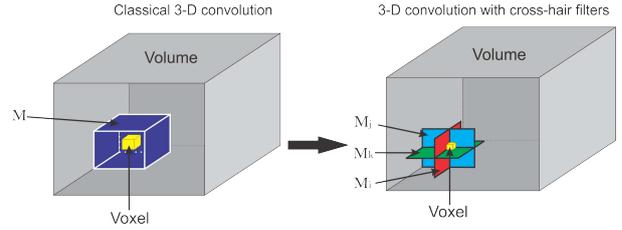

Fig. 2. Graphical representation of cross-hair filters for 3-D convolutional operation. Left: A classical 3-D convolution with filter $M$. Right: Cross-hair 3-D convolutional with 2-D filter stack $M_i, M_j, M_k$.

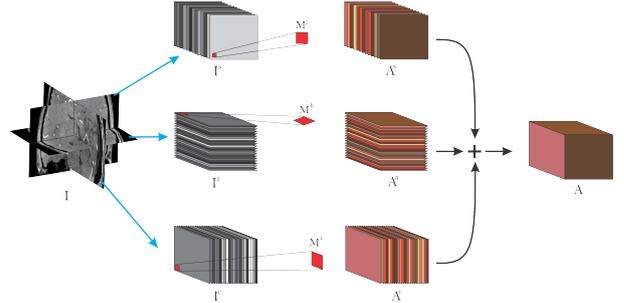

Fig. 3. Pictorial view of efficient implementation of cross-hair filters. Grayscaled stacks refer to input to the layer, red shaped squares refer to 2-D kernels used for each plane. Brown colored slices refer to extracted features after convolution operations and + symbol refers to matrix addition.

*a) Efficient Implementation:* In Eq. (2), we presented our 2-D crosshair filters. However, applying (2) independently for each voxel (as defined in Eq. (2)) leads to a redundant use of memory. More precisely, voxels close to each other share some neigbourhood information and making multiple copies of it is not memory efficient. To this end we present an efficient implementation below (Fig. 3). Consider $I$ as defined in Eq. (1) and let us extract the sagital, coronal, and axial planes as $I^s, I^c$, and $I^a$ respectively. By application of Eqs (1), and (2), we have a final implementation as follows:

$$I \diamond M = A = \beta_c A^c + \beta_s A^s + \beta_a A^a, \quad (4)$$
$$A^c = I^c \ast\ast M^i, \quad A^s = I^s \ast\ast M^j, \quad A^a = I^a \ast\ast M^k,$$

where $\ast\ast$ refers to a 2-D convolution along the first and second axes of the left matrix over all slices in the third axis. $\beta_c$, $\beta_s$, and $\beta_a$ are weights to control the input of the planes towards the final sum, for example, in the case of different spatial resolutions of the planes (we use $\beta_c = \beta_s = \beta_a = 1$ in our experiments) and $\diamond$ refers to our crosshair filter operation. This implementation is efficient in the sense that it makes use of one volume at a time instead of copies of the volume in memory where voxels share the same neighbourhood. In other words, we still have only one volume in memory but rather rotate the kernels to match the slices in the different orientaions. This lowers the memory requirements during training and inference, allowing to train on more data with little memory.

*b) 2.5-D Networks vs. 3-D Networks with Cross-hair Filters:* Its important to discuss the difference between existing 2.5-D networks and our proposed cross-hair filters. Given a 3-D task (e.g. vessel segmentation in 3-D volume), a 2.5-D based network handles the task by considering one 2-D slice at a time. More precisely, the network takes a 2-D slice (sometimes with few neighbouring slices) as input and classifies all pixels in this slice. This is repeated for each slice in the volume and the final results from the slices are fused again to form the 3-D result. On the architecture level, 2.5-D networks are 2-D networks with a preprocessing method for extracting 2-D slices and a postprocessing method for fusing 2-D results into a 3-D volume. We note that the predictions of 2.5-D networks are solely based on 2-D context information. Examples of 2.5-D networks are the implementation of UNet in [5] used for liver and lesion segmentation tasks in CT volumetric dataset, and the network architecture of [40] for annotation of lumbar vertebrae. Extensions of this approach may include a preprocessing stage where several 2-D planes are extracted and used as input channels to the 2-D network [2], [3].

On the other hand, 3-D networks based on our proposed cross-hair filters take the whole 3-D volume as input, and at each layer in the network we apply the convolutional operator discussed in Section III-A. Therefore, our filters make use of 3-D context information at each convolutional layer and do not require specific preprocessing or post processing. Our proposed method differs from classical 3-D networks in the sense that it uses less parameters and memory since it does not use full 3-D convolutions. However, it is worth noting that our filters scale exactly the same as 2.5-D (i.e. in only two directions) with respect to changes in filter and volume sizes. More precisely, given a square or cubic filter of size $k$, we have $k^2$ parameters in a 2.5-D network and $3k^2$ in our cross-hair filter based network. Increasing the filter size by a factor of $r$ will scale up as $k+r$ quadratically in both situations (i.e. $(k+r)^2$ for 2.5-D and $3(k+r)^2$ in cross-hair filter case) as compared to full 3-D networks where the parameter size scales as a cube of $k+r$.

In summary, unlike the existing 2.5-D ideas where 2-D planes are extracted at a pre-processing stage and used as input channels to a 2-D network architecture, our cross-hair filters are implemented on a layer level which help retain the 3-D information throughout the network making it a preferred option when detecting curvilinear objects in 3-D.

### B. Extreme Class Balancing with Stable Weights

We now discuss the problem of 'extreme' class imbalance and introduce a new cost function that is capable of dealing with this problem. Often in medical image analysis, the object of interest (e.g. vessel, tumor etc.) accounts for a minority of the total voxels of the image. The objects of interest in the datasets used in this work account for less than 2.5% of the voxels (the different datasets are described in Section IV-A). A standard cross entropy loss function is given by

$$\mathcal{L}(\mathbf{W}) = -\frac{1}{N} \sum_{j=1}^{N} y_j \log P(y_j = 1|X;\mathbf{W})$$
$$+ (1-y_j)\log[1 - P(y_j = 1|X;\mathbf{W})], \quad (5)$$
$$\mathcal{L}(\mathbf{W}) = -\frac{1}{N}\left(\sum_{j \in Y_+}\log P(y_j=1|X;\mathbf{W}) + \sum_{j \in Y_-}\log P(y_j=0|X;\mathbf{W})\right),$$

where $N$ is the total number of examples, $P$ is the probability of obtaining the ground truth label given the data $X$ and network weights $\mathbf{W}$, $y_j$ is the label for the $j$th example, $X$ is the feature set, $W$ is the set of parameters of the network, $Y_+$ is the set of positive labels, and $Y_-$ is the set of negative (background) labels. Using this cost function with extreme class imbalance between $Y_-$ and $Y_+$ could cause the training process to be biased towards detecting background voxels at the expense of the object of interest. This normally results in predictions with high precision against low recall. To remedy this problem, [41] proposed a biased sampling loss function for training multiscale convolutional neural networks for a contour detection task. This loss function introduced additional trade-off parameters and it samples twice more edge patches than non-edge ones for positive cost-sensitive finetuning, and vice versa, for negative cost-sensitive finetuning. Based on this, [42]

proposed a class-balancing cross entropy loss function of the form

$$\begin{aligned}\mathcal{L}(\mathbf{W}) &= -\beta \sum_{j \in Y_+} \log P(y_j = 1 | X; \mathbf{W}) \\ &- (1 - \beta) \sum_{j \in Y_-} \log P(y_j = 0 | X; \mathbf{W}),\end{aligned} \quad (6)$$

where $\mathbf{W}$ denotes the standard set of parameters of the network, which are trained with backpropagation and $\beta$ and $1 - \beta$ are the class weighting multipliers, which are calculated as $\beta = \frac{|Y_-|}{|Y|}$, $1 - \beta = \frac{|Y_+|}{|Y|}$. $P(.)$ is the probability from the final layer of the network, and $Y_+$ and $Y_-$ are the set of positive and negative class labels respectively.

*a) Challenges from Numerical Instability and High False Positive Rate:* The idea of giving more weight to the cost associated with the class with the lowest count from equation (6), has been used in other recent works [5], [43]–[45]. However, our experiments (in Section IV-D) show that the above loss function raises two main challenges.

First, there is the problem of numerical instability. The gradient computation is numerically unstable for very big training sets due to the high values taken by the loss. More precisely, there is a factor of $\frac{1}{N}$, that scales the final sum to the mean cost in the standard cross-entropy loss function in Eq. (5). This factor ensures that the gradients are stable irrespective of the size of the training data $N$. However, in Eq. (6), the weights $\beta$ and $1 - \beta$ do not scale the cost to the mean value. For high number of data points $|Y|$ (which is usually the case of voxel-wise tasks), the sums explode leading to numerical instability. For example, given a perfectly balanced data, we have $\beta = 1 - \beta = 0.5$, irrespective of the number of data points $|Y|$. Thus, increasing the size of the dataset (batch size) has no effect on the weights ($\beta$) but increases the number of elements in the summation, causing the computations to be unstable.

Second, there are challenges from high false positive rate. A high rate of false positives leading to high recall values is observed during training and at test time. This is caused by the fact that in most cases the object of interest accounts for less than 5% of the total voxels (about 2.5 % in our case). Therefore, we have a situation where $1 - \beta < 0.05$, which implies that wrongly predicting 95 background voxels as foreground is less penalized in the loss than predicting 5 foreground voxels as background. This leads to high false positive rate and, hence, high recall values.

*b) A New 'Extreme' Class Balancing Function:* To address the challenges discussed above, we introduce different weighting ratios and an additional factor to take care of the high false positive rate; and define:

$$\begin{aligned}\mathcal{L}(\mathbf{W}) &= \mathcal{L}_1(\mathbf{W}) + \mathcal{L}_2(\mathbf{W}) \quad (7) \\ \mathcal{L}_1(\mathbf{W}) &= -\frac{1}{|Y_+|} \sum_{j \in Y_+} \log P(y_j = 1 | X; \mathbf{W}) \\ &\quad -\frac{1}{|Y_-|} \sum_{j \in Y_-} \log P(y_j = 0 | X; \mathbf{W}) \\ \mathcal{L}_2(\mathbf{W}) &= -\frac{\gamma_1}{|Y_+|} \sum_{j \in Y_{f+}} \log P(y_j = 0 | X; \mathbf{W})\end{aligned}$$

$$\begin{aligned}&\quad -\frac{\gamma_2}{|Y_-|} \sum_{j \in Y_{f-}} \log P(y_j = 1 | X; \mathbf{W}) \\ \gamma_1 &= 0.5 + \frac{1}{|Y_{f+}|} \sum_{j \in Y_{f+}} |P(y_j = 0 | X; \mathbf{W}) - 0.5| \\ \gamma_2 &= 0.5 + \frac{1}{|Y_{f-}|} \sum_{j \in Y_{f-}} |P(y_j = 1 | X; \mathbf{W}) - 0.5|\end{aligned}$$

where $Y_{f+}$ and $Y_{f-}$ are the set of false positive and false negative predictions respectively and $|.|$ is the cardinality operator which measures the number of elements in the set. $\mathcal{L}_1$ is a more numerically stable version of Eq. 6 since it computes the voxel-wise, cost which scales well with the size of the dataset or batch. But the ratio of $\beta$ to $1 - \beta$ is maintained as desired. $\mathcal{L}_2$ (FP Rate Correction) is introduced to penalize the network for false predictions. However, we do not want to give false positive ($Y_{f+}$) and false negatives ($Y_{f-}$) the same weight as total predictions ($Y_+, Y_-$), since we will end up with a loss function without any class balancing because the weights will offset each other. Therefore, we introduce $\gamma_1$ and $\gamma_2$, which depend on the mean absolute distance of the wrong predicted probabilities from 0.5 (the value can be changed to suit the task). This allows us to penalize false predictions, which are very far from the central point (0.5). The false predictions ($Y_{f+}, Y_{f-}$) are obtained through a 0.5 probability threshold. Experimental results from application of FP rate correction can be found in Section **??**.

### C. Synthetic Data for Transfer Learning

To generate synthetic data, we follow the method of [7] which implements a simulator of a vascular tree that follows a generative process inspired by the biology of angiogenesis. This approach, described in [7], has initially been developed to complement missing elements of a vascular tree, a common problem in $\mu$CT imaging of the vascular bed [8]. We now use this generator to simulate physiologically plausible vascular trees that we can use in training our CNN algorithms. The simulator by Schneider et al. [7], [8] considers the mutual interplay of arterial oxygen ($O_2$) supply and vascular endothelial growth factor (VEGF) secreted by ischemic cells to achieve physiologically plausible results. Each vessel segment is modeled as a rigid cylindrical tube with radius $r$ and length $l$. It is represented by a single directed edge connecting two nodes. Semantically, this gives rise to four different types of nodes, namely root, leaf, bifurcation, and inter nodes. Each node is uniquely identified by the 3-D coordinate $\vec{P} = (x, y, z)^T$. Combining this with connectivity information, fully captures the geometry of the approximated vasculature. The radius of parent bifurcation branch $r_p$, and the radius of left ($r_l$) and right ($r_r$) daughter branches are related by a bifurcation law (also known as Murray's law) given by $r_p^\gamma = r_l^\gamma + r_r^\gamma$, where $\gamma$ is the bifurcation exponent. Our simulator enforces the Murray's law during the tree generation process. Further constraints, $\cos(\phi_l) = \frac{r_p^4 + r_l^4 - r_r^4}{2 r_p^2 r_l^2}$ and $\cos(\phi_r) = \frac{r_p^4 + r_r^4 - r_l^4}{2 r_p^2 r_r^2}$ are placed on the bifurcation angles of the left ($\phi_l$) and right ($\phi_r$) vessel extension elements respectively. This corresponds corresponds to the optimal position of the branching point $\vec{P}_b$ with respect to a minimum volume principle, another constraint enforced in the simulator from [7], [8]. The tree generation model and the bifurcation configuration is shown in Fig. 4. Detailed description of generated data is given in Section IV-A.

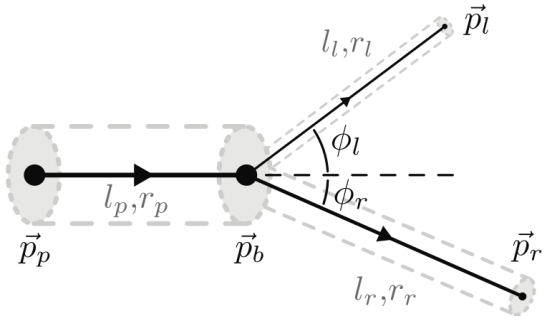

Fig. 4. A representation of the constrained bifurcation configuration, as presented in [7], where $l_p$, $l_r$, and $l_l$ are the length of the parent, right daughter, and left daughter segments, respectively. $P_r$ and $P_l$ are the right and left daughter nodes, respectively.

The output of the generation process is a tree with information on the 3-D position $\vec{P}$ of the nodes, their type (root, bifurcation, inter, leaf), and connectivity information, which includes the edge $E_{ij}$ between two nodes $N_i$ and $N_j$, and its radius $R_{ij}$. We reconstruct a 3-D volumetric data from this abstracted network description by modeling each vessel segment as a cylinder in 3-D space. We simulate different background and foreground intensity patterns with different signal-to-noise ratios.

## IV. EXPERIMENTS, RESULTS AND DISCUSSION

### A. Datasets

In this work, we use three different datasets to train and test the networks. In all three data sets, the test cases are kept apart from the training data and are used only for testing purposes. The datasets can be downloaded for public research from the paper's github page at https://github.com/giesekow/deepvesselnet/wiki/Datasets.

*a) Synthetic Dataset:* Training convolutional networks from scratch typically requires significant amounts of training data. However, assembling a properly labeled dataset of 3-D curvilinear structures, such as vessels and vessel features, takes a lot of human effort and time, which turns out to be the bottleneck for most medical applications. To overcome this problem, we generate synthetic data based on the method proposed in [7], [8]. A brief description of this process has already been presented in Section III-C. In the arterial tree generation experiment, the parameters in Table 1 of [7] are used. We use the default (underlined) values for all model parameters. We initialize the processes with different random seeds and scale the resulting vessel sizes in voxels to match the sizes of vessels in clinical datasets. Vessel intensities are randomly chosen in the interval $[128, 255]$ and non-vessel intensities are chosen from the interval $[0 - 100]$. Gaussian noise is then applied to the generated volume randomly changing the mean (i.e. in the range $[-5, 5]$) and the standard deviation (i.e. in the range $[-15, 30]$) for each volume. We generate 136 volumes of size $325 \times 304 \times 600$ with corresponding labels for vessel segmentation, centerlines, and bifurcation detection. Vessel, centerline and bifurcation labels occupy 2.1%, 0.2% and 0.05% of total intensities respectively, further highlighting the problem of class imbalance. Twenty volumes out of the 136 is used as a test set and the remaining volumes are used for pretraining our networks in the various tasks at hand. An example of the synthetic dataset can be found in Fig. 5(c). The synthetic dataset with ground truth labels for vessel, centerlines and bifurcation is available at the link provided above for download and for public use.

*b) Clinical MRA Time-of-Flight (TOF) Dataset:* To finetune and test our network architectures on real data, we obtain 40 volumes of clinical TOF MRA of the human brain, 20 of which are fully annotated and the remaining 20 partially annotated using the method proposed by [24]. Each volume has a size of $580 \times 640 \times 136$ and spacial resolution of $0.3125mm \times 0.3125mm \times 0.6mm$ on the coronal, sagittal, and axial axes respectively. We select 15 out of the 20 fully annotated volumes for testing and use the remaining five as a validation set. We also correct the 20 partially annotated volumes by manually verifying some of the background and foreground voxels. This leads to three labels, which are true foreground (verified foreground), true background (verified background), and the third class, which represent the remaining voxels not verified. In our later experiments, we use the true foreground and background labels to finetune our networks. This approach helps in avoiding any uncertainty with respect to using the partially annotated data for finetuning of the network. Image intensity ranges were scaled with a quadratic function to enhance bright structures and normalized to a standard range after clipping high intensities. A sample volume of the TOF MRA dataset can be found in Fig. 5(a). This dataset is also available at the link provided above for further research.

*c) Synchrotron Radiation X-ray Tomographic Microscopy ($\mu$CTA):* A 3-D volume of size $2048 \times 2048 \times 2740$ and spacial resolution $0.7mm \times 0.7mm \times 0.7mm$ is obtained from synchrotron radiation X-ray tomographic microscopy of a rat brain. From this large volume, we extract a dataset of 20 non-overlaping volumes of size $256 \times 256 \times 256$, which were segmented using the method proposed by [27], and use them in our later experiments to finetune the networks. To create a test set, we manually annotate 52 slices in 4 other volumes different from the 20 volumes above (208 slices in total). As with the clinical MRA data, image intensity ranges for the $\mu$CTA were also scaled with a quadratic function to enhance bright structures and normalized to a standard range after clipping high intensities. Detailed description of the $\mu$CTA data can be found in [46], and a sample volume is presented in Fig. 5(b).

### B. Network Architecture and Implementations

In this study we focus on the use of artificial neural networks for the tasks of vessel segmentation, centerline prediction and bifurcation detection. Different variants of state-of-the-art Fully Convolutional Neural Network have been presented for medical image segmentation [5], [40], [43]–[45], [47],

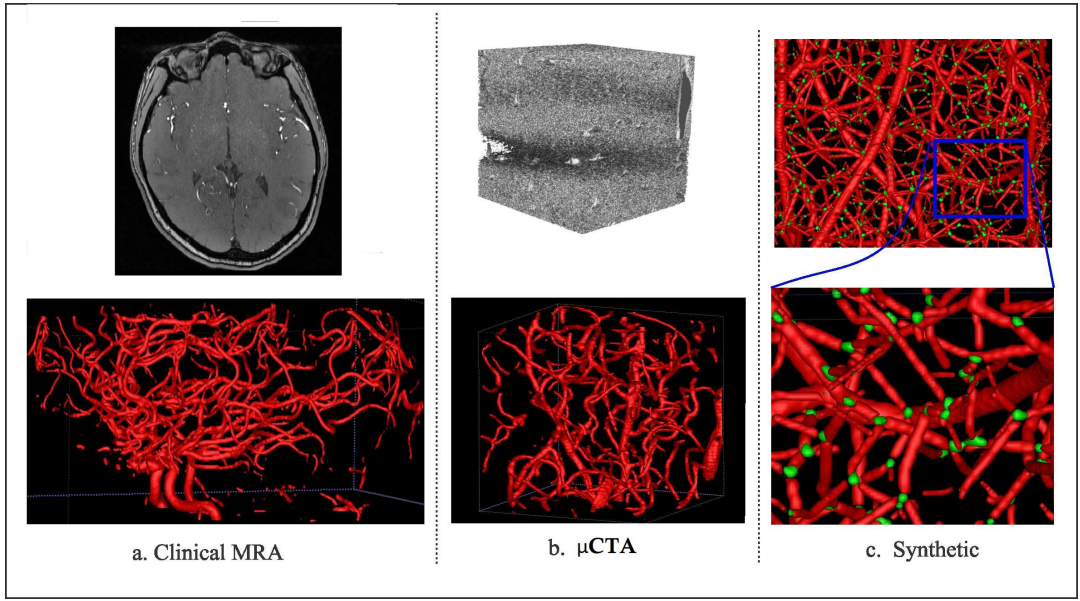

Fig. 5. Sample of datasets used in our experiments with the corresponding ground truth segmentations

[48]. Most of these architectures were based on the popular idea of convolutional-deconvolutional network which applies down-sampling at the earlier layers of the network and then reconstruct the volume at the later layers through up-sampling. This may be a bad choice given that the vascular tree tasks, especially centerline prediction and bifurcation detection, require fine details at the voxel level which can easily be lost though down-sampling. We therefore use a fully convolutional network (FCN) without down-sampling and up-sampling layers as a preferred architecture to test the performance of DeepVesselNet discussed in sections III-A, III-B and III-C. Nonetheless we also implement DeepVesselNet with popular convolutional-deconvolutional architectures to systematically study the effect of cross-hair kernel, as well as training behaviour and generalization. Python implementation of our cross-hair filters and all other codes used in our experiments is available on Github at https://github.com/giesekow/deepvesselnet for public use.

*a) DeepVesselNet-FCN:* We construct a Fully Convolutional Network FCN with four convolutional layers and a sigmoid classification layer. In this implementation, we do not use down-sampling and up-sampling layers and we carry out the convolutions in a way that the output image is of the same size as the input image by zero-padding. The removal of the down-sampling and up-sampling layers is motivated by the fact that the tasks (vessel segmentation, centerline prediction, and bifurcation detection) involve fine detailed voxel sized objects and down-sampling has an effect of averaging over voxels which causes these fine details to be lost. The alternative max-intensity pooling can easily change the voxel position of the maximum intensity later in the up-sampling stage of the network. With DeepVesselNet-FCN implementation, we have a very simple 5-layer fully-convolutional network, which takes a volume of arbitrary size and outputs a segmentation map of the same size. For the network structure and a description of the parameters, see Fig. 6.

*b) DeepVesselNet-VNet and DeepVesselNet-Unet:* To analyse the properties of our proposed cross-hair filters, we implement two alternative convolutional-deconvolutional architectures – VNet [47] and 3D UNet [49] – and replace all 3-D convolutions with our proposed cross-hair filters discussed in section III-A to obtain DeepVesselNet-VNet and DeepVesselNet-UNet respectively. By comparing the parameter size and execution time of DeepVesselNet-VNet and DeepVesselNet-UNet to the original VNet and 3D UNet implementations, we can evaluate the improvement in memory usage as well as the gain in speed that cross-hair filters offer. We also use these implementations to test whether gain in speed and memory consumption have a significant effect on prediction accuracy. Finally, DeepVesselNet-VNet and DeepVessel-UNet architectures include sub-sampling (down-sampling and up-sampling) layers. By comparing these two architecture with DeepVesselNet-FCN we can evaluate the relevance of sub-sampling when handling segmentation of fine structures like vessels and their centerlines and bifurcations.

*c) Network Configuration, Initialization, and Training:* We use the above described architecture to implement three binary networks for vessel segmentation, centerline prediction, and bifurcation detection. Network parameters are randomly initialized, according to the method proposed in [50], by sampling from a uniform distribution in the interval $(-\frac{1}{\sqrt{k_x k_y k_z}}, \frac{1}{\sqrt{k_x k_y k_z}})$ where $(k_x \times k_y \times k_z)$ is the size of the given kernel in a particular layer. For each volume in our training set, we extract non-overlapping boxes of size $(64 \times 64 \times 64)$ covering the whole volume and then feed them through the network for the finetuning of parameters. The box extraction is only done at training time to enable fast training and efficient use of computation memory, this is however not needed after our convolutional kernels are trained since full volumes can be used at test time. We train the network using a stochastic gradient descent optimizer without regularization.

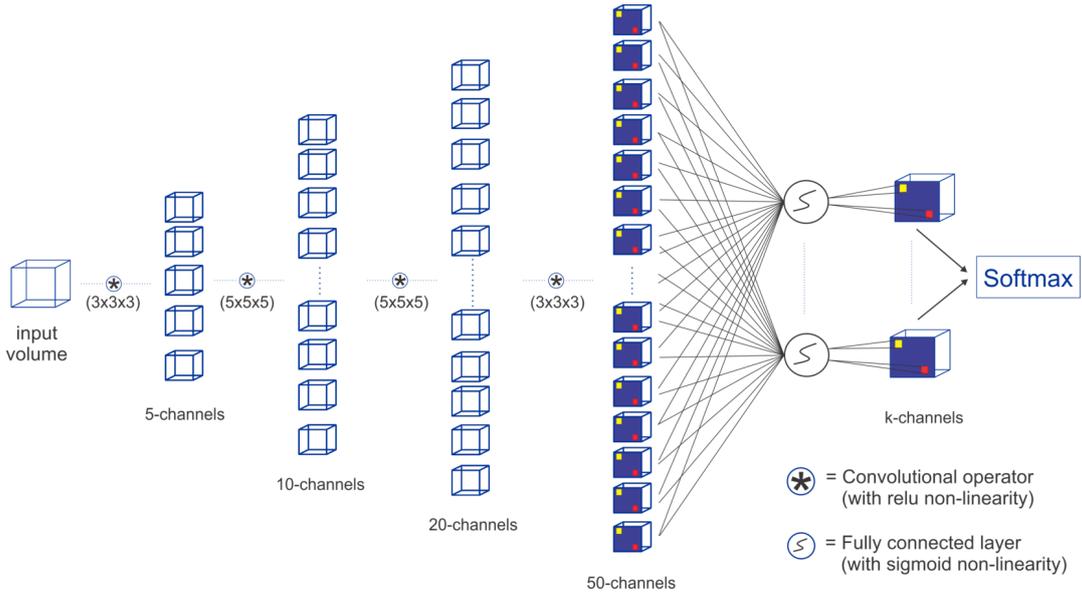

Fig. 6. Our proposed DeepVesselNet-FCN architecture implementation with crosshair filters.

During pretraining, we use a learning rate of 0.01 and decay of 0.99, which is applied after every 200 iterations for all network architectures. For finetuning, we use a learning rate of 0.001 and a decay of 0.99 applied after every 200 iterations. We implement our algorithm using the THEANO [51] Python framework and train on a machine with 64GB of RAM and Nvidia TITAN X 12GB GPU.

### C. Evaluating the DeepVesselNet Components

Prior to evaluating the performance of DeepVesselNet, we conducted a series of experiments to test the components of DeepVesselNet which includes fast cross-hair filters, the FP rate correction, and pretraining on synthetic data. These experiments and their results are discussed in this Subsection.

*a) Fast Cross-hair Filters:* To investigate the usefulness of cross-hair filters in DeepVesselNet, we experiment with full 3-D versions of DeepVesselNet and evaluate the effect on performance based on three main criteria - memory footprint based on number of parameters, computational speed based on execution time, and prediction accuracy based on Dice score. Table I shows the number of parameters in the various architectures and the execution times in the three datasets. Comparing DeepVesselNet-VNet and DeepVesselNet-UNet with their 3-D versions (VNet and UNet), we find more than 27% (16.56m vs 22.89m and 4.45m vs 7.41m respectively) reduction in memory footprint. Also, the execution time in Table I shows that cross-hair filters improve the computational speed of DeepVesselNet-VNet and DeepVesselNet-UNet by more than 23% over VNet and UNet respectively in both synthetic and clinical MRA datasets. DeepVesselNet-FCN uses very low (only 0.05m) number of parameters as compared to the other architectures due to the absence of sub-sampling layers. Scores in Table I are obtained using kernels of size $3 \times 3 \times 3$ and $5 \times 5 \times 5$, and the benefits of using sparse cross-hair filter will be even more profound with larger kernel sizes and volume sizes. Evaluation of cross-hair filters in terms of prediction accuracy is discussed in Section IV-D.

*b) The FP Rate Correction Loss Function ($\mathcal{L}_1 + \mathcal{L}_2$):* To test the effect of FP rate correction loss function discussed in section III-B, we train the DeepVesselNet-FCN architecture on a sub-sample of four clinical MRA volumes from scratch, with and without FP rate correction described in equation (7). We train for 5000 iterations and record the ratio of precision to recall every 5 iterations using a threshold of 0.5 on the probability maps. A plot of the precision-recall ratio during training without FP rate correction ($\mathcal{L}_1$ Only) and with FP rate correction ($\mathcal{L}_1 + \mathcal{L}_2$) is presented in Fig. 7. The results of this experiments suggest that training with both factors in the loss function, as proposed in Section III-B, keeps a better balance between precision and recall (i.e. a ratio closer to 1.0) than without the second factor. A balanced precision-recall ratio implies that the training process is not bias towards the background or the foreground. This helps prevent over-segmentation, which normally occurs as a result of the introduction of the class balancing.

*c) Pretraining on Synthetic Data:* We assess the usefulness of transfer learning with synthetic data by comparing the training convergence speed, and various other scores that we obtain when we pretrain DeepVesselNet-FCN on synthetic data and finetune on the clinical MRA dataset, compared to training DeepVesselNet-FCN from scratch on the clinical MRA. For this experiment, we only consider the vessel segmentation task, as no annotated clinical data is available for centerline and bifurcation tasks. Results of this experiment are reported in Table II. We achieve a Dice score of 86.39% for training from scratch without pretraining on synthetic data and 86.68% when pretraining on synthetic data. This shows that training from scratch or pretraining on synthetic data does not make much difference regarding the accuracy of the results. However, training from scratch requires about 600 iterations more than pretraining on synthetic data for the network to converge (i.e. 50% more longer).

TABLE I
NUMBER OF CONVOLUTIONAL PARAMETERS IN THE NETWORKS USED IN OUR EXPERIMENTS. FOR THE PURPOSE OF COMPARISON, THE NUMBER OF PARAMETERS STATED HERE REFERS TO ONLY THE CONVOLUTIONAL LAYERS IN THE VARIOUS ARCHITECTURES. EX. TIME REFERS TO THE AVERAGE TIME IN SECONDS REQUIRED TO PROCESS ONE VOLUME IN THE SYTHETIC AND MRA TOF DATASETS.

| Architecture | params (millions) | Ex. Time Synthetic | Ex. Time TOF MRA | Ex. Time $\mu$CTA |
| --- | --- | --- | --- | --- |
| DeepVesselNet-FCN | 0.05 | 13s | 13s | 4s |
| DeepVesselNet-VNet | 16.56 | 17s | 20s | 7s |
| DeepVesselNet-UNet | 4.45 | 13s | 14s | 4s |
| VNet | 22.89 | 23s | 26s | 11s |
| UNet | 7.41 | 17s | 19s | 6s |

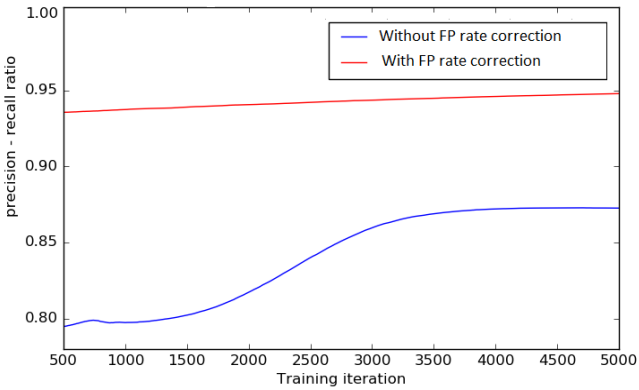

Fig. 7. Precision - recall ratio during training, with FP rate correction and without FP rate correction in the loss function, on four selected clinical MRA volumes. A balanced precision-recall ratio (i.e. close to 1) implies that we obtain the FP rate correction we propose in the work and the training process is not bias towards the background or the foreground.

TABLE II
RESULTS FROM PRETRAINING DEEPVESSELNET-FCN ON SYNTHETIC DATA AND FINETUNING WITH THE TRAINING SET FROM THE CLINICAL MRA VS. TRAINING DEEPVESSELNET-FCN FROM SCRATCH ON CLINICAL MRA. ITERATIONS REFERS TO TRAINING ITERATIONS REQUIRED FOR THE NETWORK TO CONVERGE. ALTHOUGH THE RESULT IN DICE SCORE ARE NOT VERY DIFFERENT, IT IS CLEAR THAT THE PRETRAINING ON SYNTHETIC DATA LEADS TO AN EARLIER CONVERGENCE OF THE NETWORK.

| Method | Precision | Recall | Dice | Iterations |
| --- | --- | --- | --- | --- |
| With pretraining | 86.44 | 86.93 | 86.68 | 1200 |
| Without pretraining | 85.87 | 86.92 | 86.39 | 1800 |

*D. Evaluating DeepVesselNet Performance*

In this subsection, we retain the best training strategy from the described experiments in Section IV-C and assess the performance of our proposed network architecture with other available methods mainly on the vessel segmentation task. As a further validation of our methodology we handle centerline prediction and bifurcation detection using the proposed architectures. Given a good vessel segmentation, centerline prediction and bifurcation detection tasks is classically handled by applying vessel skeletonization as a post processing step and a search of the resulting graph. Our aim in applying our architecture to handle these tasks is not to show superiority over the existing vessel skeletonization methods but it is to serve as a further verification of the effects of our described methodology and to offer a complementary way of obtaining centerlines and bifurcations, for example, to increase the robustness of the processing pipeline when fusing results of complementary approaches. The details of these experiments, results and discussion are given below.

*a) Vessel Segmentation:* We pretrain DeepVesselNet-(FCN, VNet, UNet) architectures on synthetic volumes for vessel segmentation and evaluate its performance on TOF MRA volumes through a transfer learning approach. We later finetune the networks with additional clinical MRA data, repeating the evaluation. Table III reports results of these tests, together with performances of competing methods. We obtain a Dice score of 81.48% for DeepVesselNet-FCN, 81.32% for DeepVessel-UNet and 80.10% for DeepVesselNet-VNet on TOF MRA test dataset with the transfer learning step, and 86.68% (DeepVesselNet-FCN), 84.36% (DeepVesselNet-UNet) as well as 84.25% (DeepVesselNet-VNet) after finetuning. This results (also box plots in Fig. 8) suggest that, with a Cox-Wilcoxon significance test p-value of less than 0.001, DeepVesselNet-FCN which does not use sub-sampling outperforms the versions of networks that use sub-sampling layers (VNet and UNet). Table III also reports results from the methods of [27] and [24] all of which are outperformed by DeepVesselNet-FCN in terms of Dice score. Comparing DeepVesselNet-VNet and VNet (84.25% vs 84.97% with a p-value of 0.04) as well as DeepVesselNet-Unet and UNet (84.36% vs 84.68 with a p-value of 0.07) on the MRA data, we find an advantage of up to 1% for the latter in terms of Dice scores. However, DeepVesselNet-VNet and DeepVesselNet-Unet have the advantage of being memory and computationally efficient as seen in Table I. These results show that cross-hair filters can be used in DeepVesselNet at a little to no cost in terms of vessel segmentation accuracy.

*b) Centerline Prediction:* For centerline prediction, we train DeepVesselNet on the synthetic dataset, test it on synthetic dataset and present visualizations on synthetic and clinical MRA datasets (see Figs. 12 and 11). The networks use the probabilistic segmentation masks from the vessel segmentation step as an input. Qualitative results are presented in Figs. 11 and 12 together with quantitative scores in Table IV. DeepVesselNet-VNet performs slightly worse than VNet in terms of the Dice score (66.96% vs. 74.82% with a p-value of 0.0001). Similar trend can be seen when we compare Dice scores of DeepVesselNet-UNet and UNet (72.10% vs 72.41% with a p-value of 0.0001). We obtain a Dice score of 79.92% for DeepVesselNet-FCN, which outperforms UNet and VNet and their corresponding DeepVesselNet variants with a significance test p-value of less than 0.0001. Here we note that the morphological operations based method of Schneider et al., which represents a state of the art method for centerline prediction, is able to obtain a higher recall than DeepVesselNet-FCN method (86.03% vs 82.35). This means that it detects more of the centerline points than DeepVesselNet-FCN. However, it suffers from lower precision (48.07% vs 77.63%) due to higher false positive rate which causes the overall performance to fall (61.68% vs 79.92% Dice score) as compared to DeepVesselNet-FCN. From the box plots in Fig. 9 it is very evident DeepVesselNet-FCN significantly outperforms all other architectures suggesting that the performance of the other architectures suffers from the use of sub-sampling layers.

TABLE III
RESULTS FOR VESSEL SEGMENTATION. TOF MRA ARE EVALUATED WITHIN THE BRAIN REGION ONLY. PRETRAINED RESULTS REFERS TO THE SCORES WE OBTAINED ON THE TEST SET AFTER PRETRAINING, AND FINETUNED RESULTS ARE SCORES AFTER FINETUNING WITH ANNOTATED DATA AVAILABLE FOR TOF-MRA.

| Dataset | Method | Prec. | Rec. | Dice |
|---|---|---|---|---|
| Synthetic | DeepVesselNet-FCN | **99.84** | **99.87** | **99.86** |
| | DeepVesselNet-VNet | 99.54 | 99.59 | 99.56 |
| | DeepVesselNet-UNet | 99.48 | 99.42 | 99.45 |
| | VNet | 99.48 | 99.50 | 99.49 |
| | UNet | 99.57 | 99.52 | 99.55 |
| | Schneider et al. | 99.47 | 99.56 | 99.52 |
| TOF MRA | DeepVesselNet-FCN (finetuned) | **86.44** | **86.93** | **86.68** |
| | DeepVesselNet-FCN (pretrained) | 82.76 | 80.25 | 81.48 |
| | DeepVesselNet-VNet (finetuned) | 85.00 | 83.51 | 84.25 |
| | DeepVesselNet-VNet (pretrained) | 83.32 | 77.12 | 80.10 |
| | DeepVesselNet-UNet (finetuned) | 83.56 | 85.18 | 84.36 |
| | DeepVesselNet-UNet (pretrained) | 83.48 | 79.27 | 81.32 |
| | VNet (finetuned) | 84.34 | 85.62 | 84.97 |
| | VNet (pretrained) | 82.41 | 75.82 | 78.98 |
| | UNet (finetuned) | 84.02 | 85.35 | 84.68 |
| | UNet (pretrained) | 83.16 | 80.23 | 81.67 |
| | Schneider et al. | 84.81 | 82.15 | 83.46 |
| | Forkert et al. | 84.99 | 73.00 | 78.57 |
| $\mu$CTA | DeepVesselNet-FCN | **96.72** | 95.82 | **96.27** |
| | DeepVesselNet-VNet | 95.83 | **96.18** | 96.01 |
| | DeepVesselNet-UNet | 95.85 | 96.06 | 95.95 |
| | VNet | 95.25 | 95.84 | 95.55 |
| | UNet | 95.27 | 95.71 | 95.49 |
| | Schneider et al. | 95.15 | 91.51 | 93.30 |

TABLE IV
RESULTS FOR CENTERLINE PREDICTION TASKS. RESULTS SUGGEST THAT ARCHITECTURES WITH SUB-SAMPLING LAYERS SUFFER FALL IN PERFORMANCE DUE TO LOSS OF FINE DETAILS WHICH IS CRUCIAL IN CENTERLINE PREDICTION.

| Method | Prec. | Rec. | Dice |
|---|---|---|---|
| DeepVesselNet-FCN | **77.63** | 82.35 | **79.92** |
| DeepVesselNet-VNet | 65.15 | 68.87 | 66.96 |
| DeepVesselNet-UNet | 71.28 | 72.95 | 72.10 |
| VNet | 76.41 | 73.30 | 74.82 |
| UNet | 71.25 | 73.61 | 72.41 |
| Schneider et al. | 48.07 | **86.03** | 61.68 |

TABLE V
RESULTS FROM BIFURCATION DETECTION EXPERIMENTS. PRECISION AND RECALL ARE MEASURED ON THE BASIS OF THE $5 \times 5 \times 5$ BLOCKS AROUND THE BIFURCATION POINTS. MEAN ERROR AND ITS CORRESPONDING STANDARD DEVIATION ARE MEASURED IN VOXELS AWAY FROM THE BIFURCATION POINTS (NOT THE $5 \times 5 \times 5$ BLOCKS).

| Method | Prec. | Rec. | Det. % | Mean Err | Err Std |
|---|---|---|---|---|---|
| DeepVesselNet-FCN | **78.80** | **92.97** | **86.87** | 0.2090 | **0.6671** |
| DeepVesselNet-VNet | 46.80 | 56.70 | 84.21 | 1.6533 | 0.9645 |
| DeepVesselNet-UNet | 29.47 | 88.41 | 85.89 | 0.6227 | 0.9380 |
| VNet | 25.50 | 68.71 | 70.29 | 1.2434 | 1.3857 |
| UNet | 32.57 | 77.81 | 71.78 | 1.2966 | 1.4000 |
| Schneider et al. | 77.18 | 85.08 | 84.30 | **0.1529** | 0.7074 |

*c) Bifurcation Detection:* For a quantitative evaluation of DeepVesselNet in bifurcation detection, we use synthetically generated data, and adopt a two-input-channels strategy. We use the vessel segmentations as one input channel and the centerline predictions as a second input channel relying on the same training and test splits as in the previous experiments. In our predictions we aim at localizing a cubic region of size $(5 \times 5 \times 5)$ around the bifurcation points, which are contained within the vessel segmentation. We evaluate the results based on a hit-or-miss criterion: a bifurcation point in the ground truth is counted as hit if a region of a cube of size $(5 \times 5 \times 5)$ centered on this point overlaps with the prediction, and counted as a miss otherwise; a hit is considered as true positive (TP) and a miss is considered as false negative (FN); a positive label in the prediction is counted as false positive (FP) if a cube of size $(5 \times 5 \times 5)$ centered on this point contains no bifurcation point in the ground truth. Qualitative results on synthetic and clinical MRA TOF are shown in Figs. 13 and 14, respectively. Results for Schneider et al. are obtained by first extracting the vessel tree and searching the graph for nodes. Then all nodes with two or more splits are treated as bifurcations – this being one of the standard methods for bifurcation extraction. In Fig. 10, we present the box plots of Dice score distributions obtained by the different architectures over our test set. Results from Table V and Fig. 10 show that DeepVesselNet-FCN performs better than the other architectures in 5 out of 6 metrics. In our experiments, it became evident that VNet tends to over-fit, possibly due to its high number of parameters. This may explain why results for VNet are worse than all other methods, also suggesting that in cases where little training data is available, the DeepVesselNet-FCN architecture may be the preferable due to low number of parameters and the absence of sub-sampling layers.

## V. SUMMARY AND CONCLUSIONS

We present DeepVesselNet, an architecture tailored to the challenges of extracting vessel networks and network features using deep learning. Our experiments in Sections IV-C and IV-D show that the cross-hair filters, which is one of the components of DeepVesselNet, performs comparably well as 3-D filters and, at the same time, improves significantly both speed and memory usage, easing an upscaling to larger data sets. Another component of DeepVesselNet, the introduction of new weights and the FP rate correction discussed in Section III-B, helps in maintaining a good balance between precision and recall during training. This turns out to be crucial for preventing over and under-segmentation problems, which are common problems in vessel segmentation. We also show from our results in Section IV-D that using sub-sampling layers in a network architecture in tasks which includes voxel-sized objects can lead to a fall in performance. Finally, we successfully demonstrated in Sections IV-C and IV-D that transfer learning of DeepVesselNet through pretraining on synthetically generated data improves segmentation and detection results, especially in situations where obtaining manually annotated data is a challenge.

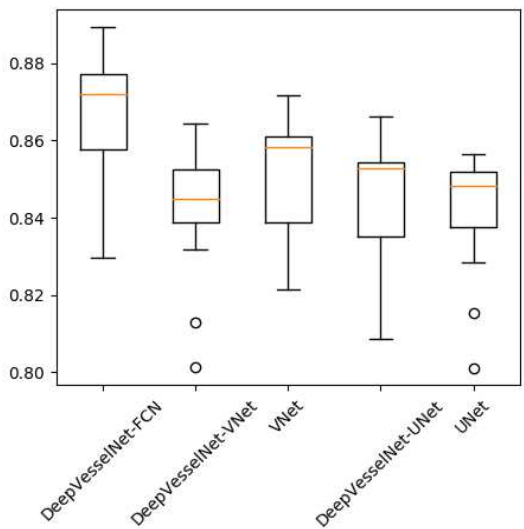

Fig. 8. Box plots of Dice scores from vessel segmentation task over our test set in the clinical MRA dataset across the deep learning architectures. It is evident that DeepVesselNet-VNet and DeepVesselNet-UNet obtain comparable results as VNet and UNet respectively. however DeepVesselNet-FCN achieves a significantly higher results as confirmed by a p-value of less than 0.001.

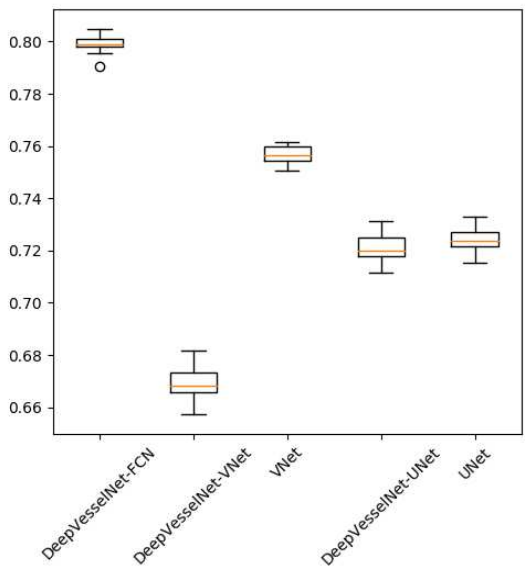

Fig. 9. Box plots of Dice scores from centerline prediction task over our test set in the synthetic dataset across the deep learning architectures. With the nature of the task involving voxel-sized structures, DeepVesselNet-FCN which does not use any form of sub-sampling significantly outperforms all other architectures.

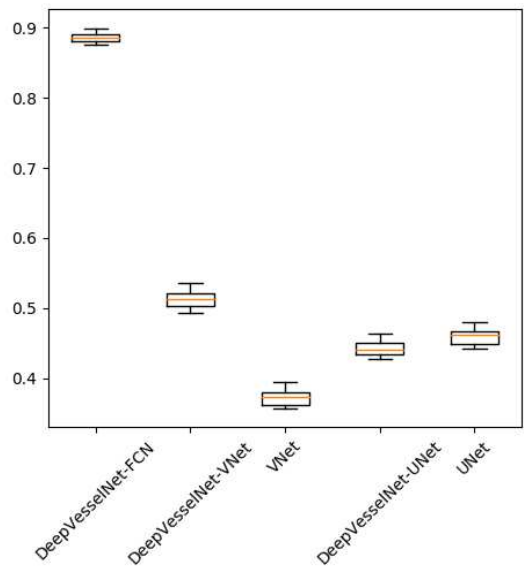

Fig. 10. Box plots of Dice scores from bifurcation detection task over our test set in the synthetic dataset across the deep learning architectures. Similar to the centerline prediction task, the nature of the task involves voxel-sized structures and DeepVesselNet-FCN again significantly outperforms all other architectures. Suggesting a negative effect of sub-sampling in voxel-sized tasks.

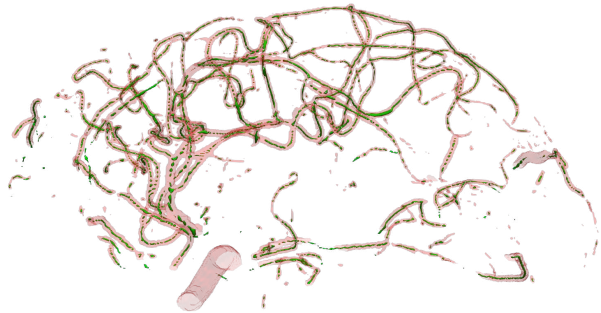

Fig. 11. Centerline prediction on MRA TOF test data using DeepVesselNet-FCN (centerline in green). Generally all vessel centerlines are detected with missing points which can be improved by finetuning on annotated MRA data or by a post-processing strategy to fill in the missing points.

As future work, we will generalize DeepVesselNet to multiclass vessel tree task, handling vessel segmentation, centerline prediction, and bifurcation detection simultaneously, rather than in three subsequent binary tasks. We also expect that network architectures tailored to our three hierarchically nested classes will improve the performance of the DeepVesselNet. For example by using a multi-level activation approach proposed in [52] or through in a single, but hierarchical approach starting from a base network for vessel segmentation, additional layers for centerline prediction, and a final set of layers for bifurcation detection.

The current implementation of cross-hair filters, network architectures and cost function are available on Github at https://github.com/giesekow/deepvesselnet. Datasets can also be downloaded from the wiki page of the same link above. Future extensions to DeepVesselNet, as well as any additional datasets will be made publicly available on Github.

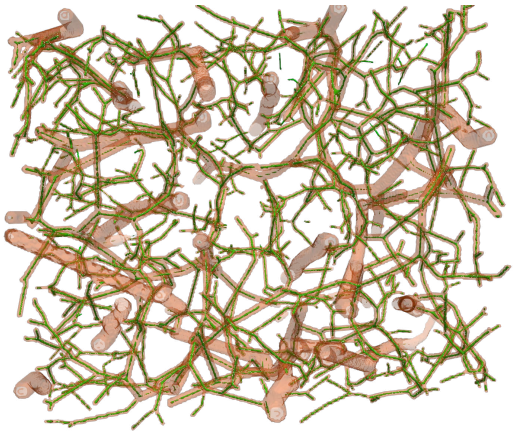

Fig. 12. Centerline prediction on synthetic test data using DeepVesselNet-FCN (centerline in green). There are more detections in smaller vessels than in larger vessels which can be explained by the network seeing more smaller vessels than bigger vessels during training. This problem can be addressed by putting more weight on centerlines in bigger vessels during training or sampling more examples from bigger vessels to balance the training data.

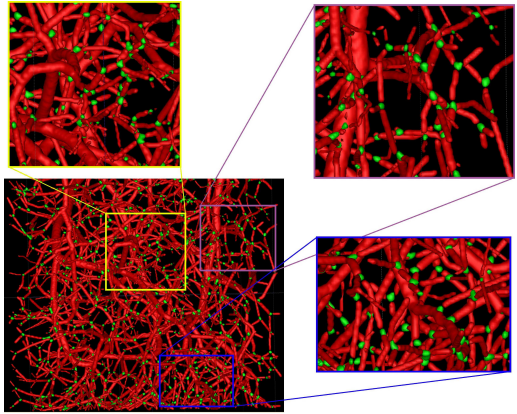

Fig. 13. Bifurcation detection on synthetic test data using DeepVesselNet-FCN (bifurcations in green). Similar to centerline prediction, bufurcation detections in smaller vessels are better than in bigger vessels which might be due to the network seeing more examples in smaller vessels than in bigger vessels during training. This problem can be remedied by putting more weight on bifurcations in bigger vessels during training or sampling more examples from bigger vessels to balance the training data.

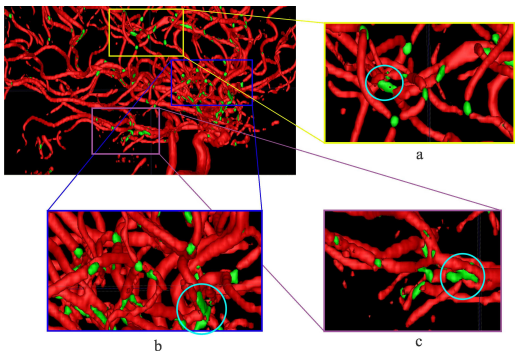

Fig. 14. Bifurcation detection in MRA TOF test data using DeepVesselNet-FCN (bifurcations in green). At regions where a lot of vessels intersect, the network predicts it as a big bifurcation, this can be seen in the circled regions in zoomed images (a, b, and c).

## REFERENCES

[1] R. Rigamonti, A. Sironi, V. Lepetit, and P. Fua, "Learning separable filters," in *The IEEE Conf. on Computer Vision and Pattern Recognition (CVPR)*, June 2013.

[2] H. R. Roth, L. Lu, A. Seff, K. M. Cherry, J. Hoffman, S. Wang, J. Liu, E. Turkbey, and R. M. Summers, "A new 2.5d representation for lymph node detection using random sets of deep convolutional neural network observations," in *Proc. MICCAI 2014*. Springer Int. Publishing, 2014, pp. 520–527.

[3] S. Liu, D. Zhang, Y. Song, H. Peng, and W. Cai, "Triple-crossing 2.5d convolutional neural network for detecting neuronal arbours in 3d microscopic images," in *Mach. Learn. in Med. Imaging*. Springer Int. Publishing, 2017, pp. 185–193.

[4] J. W. Grzymala-Busse, L. K. Goodwin, W. J. Grzymala-Busse, and X. Zheng, "An approach to imbalanced data sets based on changing rule strength," in *Rough-Neural Computing: Techniques for Computing with Words*, S. K. Pal, L. Polkowski, and A. Skowron, Eds. Berlin, Heidelberg: Springer, 2004, pp. 543–553.

[5] P. F. Christ, M. E. A. Elshaer, F. Ettlinger, S. Tatavarty, M. Bickel, P. Bilic, M. Rempfler, M. Armbruster, F. Hofmann, M. D'Anastasi, W. H. Sommer, S.-A. Ahmadi, and B. H. Menze, "Automatic liver and lesion segmentation in ct using cascaded fully convolutional neural networks and 3d conditional random fields," in *Proc. MICCAI 2016*. Cham: Springer International Publishing, 2016, pp. 415–423.

[6] G. Haixiang, L. Yijing, J. Shang, G. Mingyun, H. Yuanyue, and G. Bing, "Learning from class-imbalanced data: Review of methods and applications," *Expert Systems with Applications*, vol. 73, pp. 220 – 239, 2017.

[7] M. Schneider, J. Reichold, B. Weber, G. Székely, and S. Hirsch, "Tissue metabolism driven arterial tree generation," *Med. Image Anal.*, pp. 1397–1414, 2012.

[8] M. Schneider, S. Hirsch, B. Weber, G. Szkely, and B. Menze, "Tgif: topological gap in-fill for vascular networks–a generative physiological modeling approach," vol. 17, pp. 89–96, 01 2014.

[9] D. Szczerba and G. Székely, "Simulating vascular systems in arbitrary anatomies," in *Proc. MICCAI 2005*. Berlin, Heidelberg: Springer, 2005, pp. 641–648.

[10] C. Kirbas and F. Quek, "A review of vessel extraction techniques and algorithms." *ACM Comput. Surv*, vol. 36, pp. 81–121, 2004.

[11] D. Lesage, E. Angelini, I. Bloch, and G. Funka-Lea, "A review of 3d vessel lumen segmentation techniques: models, features and extraction schemes." *Med. Image Anal.*, vol. 13(6), p. 819845, 2009.

[12] M. E. Martínez-Pérez, A. D. Hughes, A. V. Stanton, S. A. Thom, A. A. Bharath, and K. H. Parker, "Retinal blood vessel segmentation by means of scale-space analysis and region growing," in *Proc. MICCAI 1999*. Berlin, Heidelberg: Springer, 1999, pp. 90–97.

[13] D. Nain, A. Yezzi, and G. Turk, "Vessel segmentation using a shape driven flow," in *Proc. MICCAI 2004*. Berlin, Heidelberg: Springer, 2004, pp. 51–59.

[14] A. C. S. Chung and J. A. Noble, "Statistical 3d vessel segmentation using a rician distribution," in *Proc. MICCAI 1999*, C. Taylor and A. Colchester, Eds. Berlin, Heidelberg: Springer, 1999, pp. 82–89.

[15] W. Liao, K. Rohr, and S. Wörz, "Globally optimal curvature-regularized fast marching for vessel segmentation," in *Proc. MICCAI 2013*. Berlin, Heidelberg: Springer, 2013, pp. 550–557.

[16] R. Moreno, C. Wang, and Ö. Smedby, "Vessel wall segmentation using implicit models and total curvature penalizers," in *Image Anal.: 18th Scandinavian Conf., Proc.* Berlin, Heidelberg: Springer, 2013, pp. 299–308.

[17] S. Young, V. Pekar, and J. Weese, "Vessel segmentation for visualization of mra with blood pool contrast agent," in *Proc. MICCAI 2001*. Berlin, Heidelberg: Springer, 2001, pp. 491–498.

[18] A. Dalca, G. Danagoulian, R. Kikinis, E. Schmidt, and P. Golland, "Segmentation of nerve bundles and ganglia in spine mri using particle filters," in *Proc. MICCAI 2011*. Berlin, Heidelberg: Springer, 2011, pp. 537–545.

[19] C. Florin, N. Paragios, and J. Williams, "Globally optimal active contours, sequential monte carlo and on-line learning for vessel segmentation," in *Computer Vision – ECCV 2006: 9th European Conf. on Computer Vision, Graz, Austria, Proc.* Berlin, Heidelberg: Springer, 2006, pp. 476–489.

[20] S. Wörz, W. J. Godinez, and K. Rohr, "Probabilistic tracking and model-based segmentation of 3d tubular structures," in *Bildverarbeitung für die Medizin 2009: Algorithmen — Systeme — Anwendungen Proc. des Workshops*. Berlin, Heidelberg: Springer, 2009, pp. 41–45.

[21] S. Wang, B. Peplinski, L. Lu, W. Zhang, J. Liu, Z. Wei, and R. M. Summers, "Sequential monte carlo tracking for marginal artery segmentation on ct angiography by multiple cue fusion," in *Proc. MICCAI 2013*. Berlin, Heidelberg: Springer, 2013, pp. 518–525.

[22] A. F. Frangi, W. J. Niessen, K. L. Vincken, and M. A. Viergever, "Multiscale vessel enhancement filtering," in *Proc. MICCAI 1998*, W. M. Wells, A. Colchester, and S. Delp, Eds. Berlin, Heidelberg: Springer, 1998, pp. 130–137.

[23] M. W. K. Law and A. C. S. Chung, "Three dimensional curvilinear structure detection using optimally oriented flux," in *Computer Vision – ECCV 2008*, D. Forsyth, P. Torr, and A. Zisserman, Eds. Berlin, Heidelberg: Springer, 2008, pp. 368–382.

[24] N. D. Forkert, A. Schmidt-Richberg, J. Fiehler, T. Illies, D. Möller, D. Säring, H. Handels, and J. Ehrhardt, "3d cerebrovascular segmentation combining fuzzy vessel enhancement and level-sets with anisotropic energy weights," *Magn. Reson. Imaging*, vol. 31, pp. 262–271, 2013.

[25] N. D. Forkert, A. Schmidt-Richberg, J. Fiehler, T. Illies, D. Möller, H. Handels, and D. Säring, "Fuzzy-based vascular structure enhancement in time-of-flight mra images for improved segmentation," *Methods of Information Medicine*, vol. 50, pp. 74–83, 2011.

[26] R. Phellan and N. D. Forkert, "Comparison of vessel enhancement algorithms applied to time-of-flight mra images for cerebrovascular segmentation," *Med. Physics*, vol. 44, no. 11, pp. 5901–5915, 2017.

[27] M. Schneider, S. Hirsch, B. Weber, G. Székely, and B. H. Menze, "Joint 3-d vessel segmentation and centerline extraction using oblique hough forests with steerable filters," *Med. Image Anal.*, vol. 19, no. 1, pp. 220–249, 2015.

[28] D. C. Ciresan, A. Giusti, L. M. Gambardella, and J. Schmidhuber, "Deep neural networks segment neuronal membranes." *electron microscopy images. In NIPS*, p. 28522860, 2012.

[29] R. Phellan, A. Peixinho, A. Falcão, and N. D. Forkert, "Vascular segmentation in tof mra images of the brain using a deep convolutional neural network," in *Intravascular Imaging and Computer Assisted Stenting, and Large-Scale Annotation of Biomedical Data and Expert Label Synthesis*. Springer Int. Publishing, 2017, pp. 39–46.

[30] M. Koziński, A. Mosinska, M. Salzmann, and P. Fua, "Learning to segment 3d linear structures using only 2d annotations," in *Medical Image Computing and Computer Assisted Intervention – MICCAI 2018*, A. F. Frangi, J. A. Schnabel, C. Davatzikos, C. Alberola-López, and G. Fichtinger, Eds. Cham: Springer International Publishing, 2018, pp. 283–291.

[31] B. Shagufta, S. A. Khan, A. Hassan, and A. Rashid, "Blood vessel segmentation and centerline extraction based on multilayered thresholding in ct images," *Proc. of the 2nd Int. Conf. on Intelligent Systems and Image Processing*, pp. 428–432, 2014.

[32] M. Maddah, A. Afzali-khusha, and H. Soltanian, "Snake modeling and distance transform approach to vascular center line extraction and quantification," *Computerized Med. Imag. and Graphics*, vol. 27 (6), pp. 503–512, 2003.

[33] D. Chen and L. D. Cohen, "Piecewise geodesics for vessel centerline extraction and boundary delineation with application to retina segmentation," in *Scale Space and Variational Methods in Computer Vision: 5th Int. Conf., SSVM 2015, Lège-Cap Ferret, France, May 31 - June 4, 2015, Proc.* Springer Int. Publishing, 2015, pp. 270–281.

[34] A. Santamaría-Pang, C. M. Colbert, P. Saggau, and I. A. Kakadiaris, "Automatic centerline extraction of irregular tubular structures using probability volumes from multiphoton imaging," in *Proc MICCAI 2007, Part II*. Berlin, Heidelberg: Springer, 2007, pp. 486–494.

[35] Y. Zheng, J. Shen, H. Tek, and G. Funka-Lea, "Model-driven centerline extraction for severely occluded major coronary arteries," in *Mach. Learn. in Medical Imaging: Third Int. Workshop, MLMI 2012, Held in Conjunction with MICCAI: Nice, France, Revised Selected Papers*. Berlin, Heidelberg: Springer, 2012, pp. 10–18.

[36] M. M. G. Macedo, C. Mekkaoui, and M. P. Jackowski, "Vessel centerline tracking in cta and mra images using hough transform," in *Progress in Pattern Recognition, Image Anal., Computer Vision, and Applications: 15th Iberoamerican Congress on Pattern Recognition, CIARP 2010, Sao Paulo, Brazil, November 8-11, 2010. Proc.* Berlin, Heidelberg: Springer, 2010, pp. 295–302.

[37] M. Rempfler, M. Schneider, G. D. Ielacqua, X. Xiao, S. R. Stock, J. Klohs, G. Székely, B. Andres, and B. H. Menze, "Reconstructing cerebrovascular networks under local physiological constraints by integer programming," *Med. Image Anal.*, pp. 86–94, 2015.

[38] T. Chaichana, Z. Sun, M. Barrett-Baxendale, and A. Nagar, "Automatic location of blood vessel bifurcations in digital eye fundus images," in *Proc. of Sixth Int. Conf. on Soft Computing for Problem Solving: SocProS 2016, Volume 2*. Singapore: Springer Singapore, 2017, pp. 332–342.

[39] Y. Zheng, D. Liu, B. Georgescu, H. Nguyen, and D. Comaniciu, "3d deep learning for efficient and robust landmark detection in volumetric data," in *Proc. MICCAI 2015*, N. Navab, J. Hornegger, W. M. Wells, and A. Frangi, Eds. Springer Int. Publishing, 2015, pp. 565–572.

[40] A. Sekuboyina, A. Valentinitsch, J. Kirschke, and B. H. Menze, "A localisation-segmentation approach for multi-label annotation of lumbar vertebrae using deep nets," *arXiv*, vol. 1703.04347, 2017.

[41] J.-J. Hwang and T.-L. Liu, "Pixel-wise deep learning for contour detection," in *ICLR*, 2015.

[42] S. Xie and Z. Tu, "Holistically-nested edge detection," in *Proc. of the IEEE Int. Conf. on computer vision*, 2015, pp. 1395–1403.

[43] K.-K. Maninis, J. Pont-Tuset, P. Arbeláez, and L. Van Gool, "Deep retinal image understanding," in *Proc. MICCAI 2016, Part II*. Springer Int. Publishing, 2016, pp. 140–148.

[44] I. Nogues, L. Lu, X. Wang, H. Roth, G. Bertasius, N. Lay, J. Shi, Y. Tsehay, and R. M. Summers, "Automatic lymph node cluster segmentation using holistically-nested neural networks and structured optimization in ct images," in *Proc. MICCAI 2016, Part II*. Springer Int. Publishing, 2016, pp. 388–397.

[45] H. R. Roth, L. Lu, A. Farag, A. Sohn, and R. M. Summers, "Spatial aggregation of holistically-nested networks for automated pancreas segmentation," in *Proc. MICCAI 2016, Part II*. Springer Int. Publishing, 2016, pp. 451–459.

[46] J. Reichold, M. Stampanoni, A. L. Keller, A. Buck, P. Jenny, and B. Weber, "Vascular graph model to simulate the cerebral blood flow in realistic vascular networks," *Journal of Cerebral Blood Flow & Metabolism*, vol. 29, no. 8, pp. 1429–1443, 2009.

[47] F. Milletari, N. Navab, and S. Ahmadi, "V-net: Fully convolutional neural networks for volumetric med. image segmentation," in *Fourth Int. Conf. on 3D Vision (3DV)*, 2016, pp. 565–571.

[48] G. Tetteh, M. Rempfler, C. Zimmer, and B. H. Menze, "Deep-fext: Deep feature extraction for vessel segmentation and centerline prediction," in *Mach. Learn. in Med. Imaging*. Springer Int. Publishing, 2017, pp. 344–352.

[49] Ö. Çiçek, A. Abdulkadir, S. S. Lienkamp, T. Brox, and O. Ronneberger, "3d u-net: Learning dense volumetric segmentation from sparse annotation," *CoRR*, vol. abs/1606.06650, 2016. [Online]. Available: http://arxiv.org/abs/1606.06650

[50] Y. Bengio and X. Glorot, "Understanding the difficulty of training deep feedforward neuralnetworks," in *Proc. of the 13th Int. Conf. on Artificial Intelligence and Statistics (AISTATS)*, vol. 9, 2010.

[51] Theano Development Team, "Theano: A Python framework for fast computation of mathematical expressions," *arXiv e-prints*, vol. abs/1605.02688, May 2016.

[52] M. Piraud, A. Sekuboyina, and B. H. Menze, "Multi-level activation for segmentation of hierarchically-nested classes," in *Computer Vision – ECCV 2018 Workshops*, L. Leal-Taixé and S. Roth, Eds. Cham: Springer International Publishing, 2019, pp. 345–353.